\pdfoutput=1
\documentclass[11pt]{article}
\usepackage[final]{acl}

\usepackage{times}
\usepackage{latexsym}
\usepackage{xcolor,colortbl}

\usepackage[T1]{fontenc}
\usepackage[utf8]{inputenc}

\usepackage{microtype}

\usepackage{inconsolata}

\usepackage{graphicx}
\usepackage{amsmath}
\usepackage{amssymb}
\usepackage{multirow}
\usepackage{multicol}
\usepackage{array}

\title{Exploring Curriculum Learning for Vision-Language Tasks: A Study on Small-Scale Multimodal Training}

\author{
 \textbf{Rohan Saha\textsuperscript{1}},
 \textbf{Abrar Fahim\textsuperscript{1}},
 \textbf{Alona Fyshe\textsuperscript{1,2}},
 \textbf{Alex Murphy\textsuperscript{1}}
\\
 \textsuperscript{1}Department of Computing Science,
 \textsuperscript{2}Department of Psychology \\ University of Alberta
\\
 {
    \{rsaha, afahim2, alona, amurphy3\}@ualberta.ca
 }
}

\begin{document}
\setlength{\parskip}{0pt}
\maketitle
\begin{abstract}
For specialized domains, there is often not a wealth of data with which to train large machine learning models. In such limited data / compute settings, various methods exist aiming to \textit{do more with less}, such as finetuning from a pretrained model, modulating difficulty levels as data are presented to a model (curriculum learning), and considering the role of model type / size. Approaches to efficient \textit{machine} learning also take inspiration from \textit{human} learning by considering use cases where machine learning systems have access to approximately the same number of words experienced by a 13 year old child (100M words). We investigate the role of 3 primary variables in a limited data regime as part of the multimodal track of the BabyLM challenge. We contrast: (i) curriculum learning, (ii), pretraining (with text-only data), (iii) model type. We modulate these variables and assess them on two types of tasks: (a) multimodal (text+image), and (b) unimodal (text-only) tasks. We find that curriculum learning benefits multimodal evaluations over non-curriclum learning models, particularly when combining text-only pretraining. On text-only tasks, curriculum learning appears to help models with smaller trainable parameter counts. We suggest possible reasons based on architectural differences and training designs as to why one might observe such results.

\end{abstract}

\section{Introduction}
\label{sec:introduction}
Recent vision-language models (VLMs) have achieved superior performance on numerous benchmark datasets (such as the Llama\footnote{\url{https://llama.meta.com/}} and Gemini models\footnote{\url{https://deepmind.google/technologies/gemini/}}), and continue advancing rapidly as models are scaled up. The number of parameters of such models is often on the order of billions. These models require multiple days of compute, and hundreds of GPUs (e.g., \citet{radford2021learningtransferablevisualmodels}), resulting in massive energy consumption \citep{luccioni_gpus}. Furthermore, to train such large models, we require massive amounts of pretraining data. For example, 70M image-text pairs were used to train the Flava foundation model \cite{singh2022flavafoundationallanguagevision}. Pretraining VLMs on such large scale data is often infeasible for independent researchers and university research labs with limited compute. \par
In contrast to \textit{machine} learning, \textit{human} learning is much more efficient, a finding which has led researchers to consider which methods might promote more \textit{human-like} learning in artificial neural networks. This was originally argued for in early work on curriculum learning \citep{bengio_2009_curriculum_learning}, citing the fact that humans do not learn from randomly sampled data, but benefit from learning over structured chunks, typically increasing in difficulty (a curriculum).

To this end, we explore the application of curriculum learning to VLMs with limited input data as part of the BabyLM challenge \cite{babylm-2024}. For the multimodal track, which contains a dataset of image-caption pairs, we take inspiration from phase-based curriculum methodology used in \citet{ayyubi2023learning}. We use Part-of-Speech (PoS) linguistic features from the captions to categorize samples into different phases, to generate a learning curriculum. However, instead of training the model only one phase at a time (as used in \citet{ayyubi2023learning}), we train the model on the current and previous phases such that the pool of data which can be sampled increases at each phase.
\par
From our experiments, we observe that:
\begin{itemize}
    \item In a limited data setting, curriculum learning can improve the performance of VLMs on certain multimodal and text-only evaluation benchmarks.
    \item Pretraining VLMs on developmentally plausible text-only data prior to adapting to multimodal data may help improve performance on some evaluation tasks, but not others.
\end{itemize}

\section{Background}
\label{sec:background}

% \textcolor{red}{Relation to smaller models (what happens when the model size is increased  - CL doesn't really matter - thus making it appropriate for smaller models (use this is as a motivation for referring to tons of previous papers in babylm 2023)) / VLMs / relation to the current babylm 2024 challenge.}

\subsection{Curriculum Learning}
\label{ref:cl_background}

Curriculum Learning (CL) takes inspiration from the learning process in humans by presenting data to a machine learning model in an easy-to-difficulty manner \cite{elmanLearningDevelopmentNeural1993, bengio_2009_curriculum_learning}. CL consists of two parts: (1) a scoring function to rank data samples based on difficulty, and (2) a pacing function, which controls the distribution of data samples presented to the model. In the standard CL implementation, the pacing function introduces can be samples in ascending order of difficulty (or decreasing difficulty in the case of anti-curriculum learning \cite{hacohen2019power, wu2021curriculawork}).
\par
While extensive research has shown that in certain cases, curriculum learning can provide performance gains in vision \cite{hacohen2019power, wang2019dynamiccurriculumlearningimbalanced, sovianyCurriculumLearningDiversity2020} and Natural Language Processing (NLP) tasks \cite{nagatsuka-etal-2021-pre, maharana-bansal-2022-curriculum, sun2023data}, in other cases, the benefit is unclear \cite{campos2021curriculum, martinezCLIMBCurriculumLearning2023, chobey-etal-2023-training, edman2023too}. Importantly, with the prevalence of vision-language models, it is crucial to understand how the application of CL modulates VLMs to work in the domain of limited data and compute.

\subsection{Curriculum Learning for Vision Language Models}
\label{sec:cl_for_vlms_background}
% Although bigger and better models with vision capabilities bring better performance, the need for greater compute has grown significantly. Thus, research in efficient pretraining strategies CL for VLMs has become ever important. \par
Some previous work has applied CL to multimodal models where the data modality consists of images and texts. \citet{srinivasan2023curriculum} showed that CL applied to a transformer model helps improve performance on zero-shot image and text retrieval tasks over a baseline CLIP model \cite{radford2021learningtransferablevisualmodels}. CL has also shown benefit in other multimodal domains, such as medical report generation \cite{liu2023competencebasedmultimodalcurriculumlearning}, image-captioning \cite{ayyubi2023learning}, and visual question answering \cite{li2020competence_vqa}. 
However, these works either rely on non vision-transformer based image encoders (such as an R-CNN), or conduct evaluation on a small set of evaluation tasks. It is also unclear whether: (i) training VLMs on image-caption data improves model performance on text-only benchmarks; (ii) how CL affects downstream performance in models with additional text pretraining compared to randomly initialized models.

\par

In this work, we present a study where we apply CL to VLMs trained on \textit{small data}. We hope to provide the research community with a better understanding of the effects of CL on popular VLMs such as the Generative Image Transformer (GIT) \cite{wang2022gitgenerativeimagetotexttransformer} and Flamingo \cite{alayrac2022flamingovisuallanguagemodel} models. Furthermore, we also explore the effect of CL on downstream model performance on various zero-shot multimodal and text-based benchmarks.

\section{Methods}
\label{sec:methods}

\subsection{Data}
\label{sec:data}

We use the dataset provided as part of the BabyLM multimodal track \cite{babylm-2024}. The data consist of 100M words in total: 50M words from varied text corpora (described in \citet{babylm-2024}) and the other 50M words are text captions taken from the Conceptual Captions \cite{sharma2018conceptual} and Localized Narratives \cite{PontTuset_eccv2020} image-caption datasets. In total, the multimodal data consists of $\sim$ 2.9M image-caption pairs. \par
One of the key experimental variables we examine is the impact of text pretraining. For multimodal models, we compare the performance of models trained on image-caption data (consisting of 50M words), starting either from a randomly initialized model or from a model pretrained on the text-only corpora mentioned above (50M words). Model variants not pretrained on the text-only corpora only use the words in the captions of the associated training images (i.e., models are trained on only 50M words and the corresponding images).

\subsection{Models}
\label{sec:framework}

We train two VLMs: (1) $GIT$ \citet{wang2022gitgenerativeimagetotexttransformer} and (2) Flamingo \cite{alayrac2022flamingovisuallanguagemodel}. We chose these models because they were selected as reference baselines provided by the BabyLM challenge \cite{babylm-2024}. Both $GIT$ and $Flamingo$ models consist of vision encoders to encode image inputs, and text decoders to generate free-form text. 

% $GIT$ projects image embeddings into the same space as the text embeddings and fuses representations via the self-attention mechanism. For $Flamingo$, cross-attention between image and caption is calculated and fused between LM blocks. $Flamingo$ accepts multiple references to images in a single sample's text captions, but applies masking such that each text token only attends to the previous image representation. In our use case, we only consider single image-caption pairs, so this property of the $Flamingo$ architecture reduces to standard cross-attention. \par

We use the default configurations for the $GIT$\footnote{https://huggingface.co/babylm/git-2024} and $Flamingo$\footnote{https://huggingface.co/babylm/flamingo-2024} models provided in the BabyLM challenge to compare the performance of our models to the baselines reported by the challenge. Following the default configurations, we use pretrained vision encoders\footnote{https://huggingface.co/facebook/dino-vitb16} for both the $GIT$ and $Flamingo$ models.
Furthermore, according to default model configurations, we update all model parameters for the $GIT$ model, but for $Flamingo$, we keep the vision encoder frozen, and update all other parameters. $GIT$ has a total of 198 million parameters (198 million trainable parameters), and $Flamingo$ has 255 million total parameters (169 million trainable parameters because of the frozen vision encoder).

\paragraph{Tokenizer:} Pretrained tokenizers are trained on data that exceed the limit imposed by the challenge. Thus, we train a new \textit{WordPiece} tokenizer (using a \texttt{bert-base-uncased} model configuration) from scratch on the text-only and caption data (100M words total). 
We use the same tokenizer for both $GIT$ and $Flamingo$ to avoid confounding model performance differences with the tokenizer choice.

\par
\subsection{Curriculum Framework}
\label{sec:curriculum_learning}
We discuss the respective implementations of the scoring and pacing functions for the curriculum learning framework below.

\paragraph{Scoring function:} A scoring function assigns a \textit{difficulty score} $k \in \mathbb{R}$ to each sample in the dataset, where a sample $x_i$ is easier than a sample $x_{i+1}$, if $k_{x_i} < k_{x_{i+1}}$. \par
Previous works have used a variety of scoring functions to measure sample difficulty, such as the loss scoring function in image classification \cite{hacohen2019power} and text classification settings \cite{xu-etal-2020-curriculum, maharana-bansal-2022-curriculum}. Relatedly, in sample-efficient pretraining of language models, average sentence rarity \cite{borazjanizadeh-2023-optimizing}, sentence length \cite{debenedetto2023byte} or other combinations of individual text statistics \cite{edman-bylinina-2023-much} have been used to rank data samples (for a comprehensive survey, see \citet{sovianyCurriculumLearningSurvey2021}).
More recently, in multimodal settings, cross-modal similarity \cite{zhang-etal-2022-cross} has been used to rank examples to improve model performance in image-captioning tasks. All in all, it must be noted that determining the difficulty of image-caption pairs is non-trivial and an active research problem.
\par

\begin{figure}[!htb]
    \centering
\includegraphics[width=1.0\linewidth]{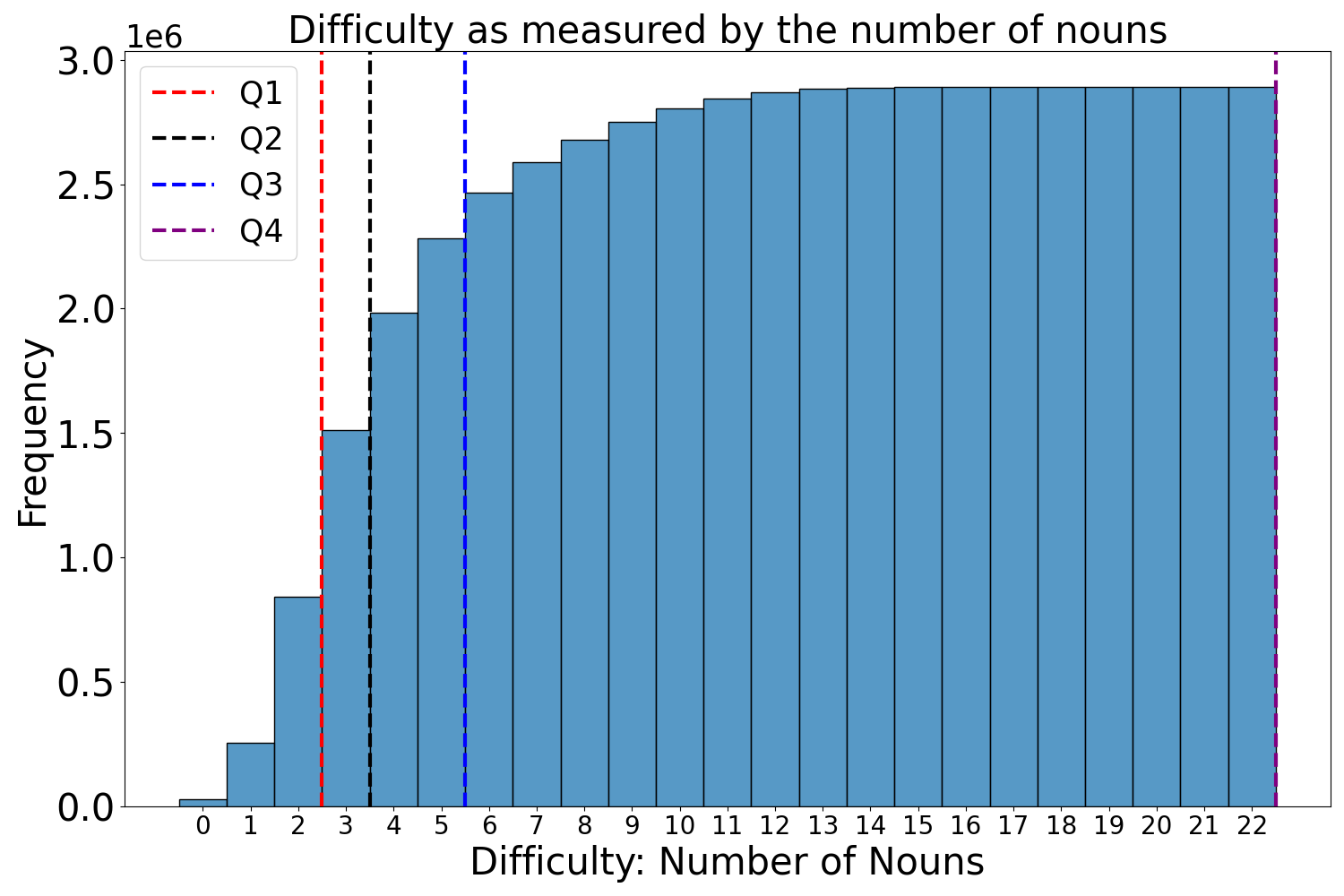}
    \caption{Cumulative distribution of scores for all the image-caption pairs. The dashed vertical lines determine each of the four quartiles, where each quartile contains the samples that belong to a specific curriculum phase.}
    \label{fig:cumulative_distribution_scores_all_caption_data}
\end{figure}

For our experiments, we explored the applicability of linguistic information such as Part-of-Speech (PoS) tags to determine difficulty of samples. We took inspiration from the scoring function used by \citet{ayyubi2023learning}, where a PoS tagger was used to count the number of nouns in the caption, as an indirect measure of the number of concepts present in the image. The number of concepts, in turn, determined the difficulty of the image-caption pair. 

As the BabyLM challenge has limits on the number of words that can be used to train systems, we trained our own PoS tagger to tag the image captions. To train the tagger, we first created a training dataset by annotating the provided text-only and caption data, with POS symbols\footnote{These are non-word elements such as NN for noun, or JJ for determiner}, using an off-the-shelf PoS tagger from NLTK \footnote{\url{https://www.nltk.org/api/nltk.tag.pos_tag.html}}. Then we used this newly created annotated training dataset to train a custom PoS tagger on the permissible limited text words. We implemented the PoS tagger using a token classification model using $BERT_{BASE}$ as the backbone model architecture. We trained the tagger for 5 epochs\footnote{We observed that 5 epochs were sufficient to achieve $\sim$97.42\% accuracy on a 10\% held out validation dataset. We then trained the tagger on all the data (train+validation).}, using a batch size of 512 and half-precision (FP16) training. \par

\paragraph{Distribution of difficulty scores:} We show the cumulative distribution of the scores assigned by the PoS scoring function in Figure \ref{fig:cumulative_distribution_scores_all_caption_data}. For images having multiple captions, we consider the maximum value of the difficulty (maximum number of nouns) amongst all the captions for that image. We use maximum difficulty to account for the most complex interpretation of the image and avoid underestimation of the difficulty value.

\paragraph{Ordering:} In our experiments, we order the samples in ascending order of difficulty, to explore the performance improvement of unimodal and multimodal models when they are trained in a manner similar to how humans acquire novel information. Although previous work has shown that a descending ordering of difficulty can be beneficial for model performance for certain tasks (e.g., \citet{maharana-bansal-2022-curriculum}), we leave this for future research given limited compute.

\paragraph{Pacing function:} A pacing function controls the rate at which samples of different training curriculum phases are presented to the model. Multiple different pacing strategies exist, such as fixed exponential pacing, step pacing for image classification \cite{hacohen2019power}, competence function \cite{platanios-etal-2019-competence} for machine translation, to name a few. \par 
For our experiments, we design a simple pacing function inspired by the phase-level pacing function \cite{ayyubi2023learning} and competence-based pacing function \cite{platanios-etal-2019-competence}. We use the quartiles from the cumulative distribution of the sample difficulty scores (Figure \ref{fig:cumulative_distribution_scores_all_caption_data}), giving us four \textit{blocks} of difficulty levels. For simplicity, we also train our model in four phases, where in each training phase $p$, we train the model on samples that have difficulty levels in the $p^{th}$ quartile. For example, in Figure \ref{fig:cumulative_distribution_scores_all_caption_data}, the first phase ($p_1$) contains samples with difficulty level $k \le 2$, the second phase contains samples with difficulty level $k \le 3$, the third phase contains samples with $k \le 5$, while the fourth phase contains all the samples in the dataset. For each training phase, we randomly sample training batches from the set of data available up to the corresponding training quartile. It must be noted with each new block, the number of available data points increases, which has an effect during training, where earlier epochs are faster (because of fewer samples) compared to later epochs. \par
This approach contrasts the phase-level curriculum learning introduced by \citet{ayyubi2023learning}, where the model is trained only on samples from a specific block, which may cause the model to focus more on samples in that specific block, while not retaining previously learned information from earlier phases. Furthermore, our pacing function has the added advantage of not requiring extensive hyperparameter tuning, such as the exponential pacing function used by \citet{hacohen2019power}, and is thus suitable for scenarios with limited computational resources.

% \subsection{Model performance in limited training time}
% \label{sec:model_performance_in_limited_training_time}
% Given limited training time, curriculum learning has been previously been found to benefit model performance \cite{wu2021curriculawork} in image-classification settings. We explore whether this phenomenon holds true for vision-language models. <TODO: incorporate this: We find that this actually does not hold true for VLMs. >

\subsection{Models Variants}
\label{sec:models}
For both $GIT$ and $Flamingo$, we train four model variants, two of which are baseline models and two are trained using curriculum learning. In each pair, we train one model only on the image-caption data starting from random initialization (except the vision encoder which is pretrained), while we first pretrain the other variant on the text-only corpus, before training on image-caption data.
\paragraph{Baselines:} For the first baseline variant, we train the model on the image-caption dataset (50M words) using standard i.i.d training. We refer to this variant with \textbf{C} (denoting that the model is trained on the image-caption data only) for both $GIT_{Baseline}$ and $Flamingo_{Baseline}$.
For the second baseline variant, we first train the model on the text-only dataset (containing 50M words) using standard i.i.d training. We then continue the training procedure on image-caption dataset (containing another 50M words) using standard i.i.d training. We refer to this variant as \textbf{T+C}, for both $GIT_{Baseline}$ and $Flamingo_{Baseline}$. \par

Our choice to also train the \textbf{T+C} model variant stems from previous work showing that exposing the model to developmentally plausible data, such as child-directed speech, before exposing it to complex data, can benefit model performance \cite{huebner-etal-2021-babyberta}. 
Thus, we explore the difference in model performance, when we first train the model on the text-only dataset, before continuing the training procedure on the image-caption data. 

% Therefore, we have four baseline models: $GIT$ (with and without text initialization) and $Flamingo$ (with and without text initialization).

\paragraph{Curriculum models:}
For curriculum variants, we use CL on the image-caption pairs because we hypothesize that applying CL on multimodal data will improve model performance. We refer to these variants trained only on the image-caption pairs as \textbf{C} under $GIT_{CL}$ and $Flamingo_{CL}$. We also train \textbf{T+C} variants of CL models, where we first pretrain the model on the text-only dataset using standard i.i.d training, and then use curriculum learning to continue the training procedure on the image-caption pairs.

To summarize, we trained four variants for each model, two of which were trained using standard training (no curriculum), and the other two were trained using curriculum learning.
For $GIT$ and $Flamingo$ baseline variants, we train the model on the image-caption only (\textbf{C}) data, and both text + image-caption (\textbf{T+C}) data. Similarly, for the curriculum variants, we train each model on, image-caption data only (\textbf{C}) data, and both text + image (\textbf{T+C}) data. 
% Where the For our experiments in this category, we only applied curriculum learning to the \textbf{C} portion of the training stage (the stage with image-caption data), while applying standard i.i.d. training on the \textbf{T} portion (stage with just text data).

% For our experiments, we only applied curriculum to the multimodal data. 

\section{Training and Evaluation Details}
\label{sec:training_and_eval_details}

\paragraph{Training Details:}  For the curriculum variants, we train the model for two epochs per each \textit{difficulty phase} (of which there are four). We used a learning rate of $1e^{-5}$, maximum token length of $50$, and 32 samples per batch \footnote{We use a batch size of $32$ when training on the image-caption data, but we use a value of 256 when pretraining the model (T+C variant) on the text-only dataset as memory requirements are lower.}, and Adam optimizer\footnote{We use default hyperparameters for Adam: $\beta_1$ = 0.9, $\beta_2$ = 0.999, eps=$1e^{-08}$, weight\_decay=0.} \cite{kingma2017adammethodstochasticoptimization}. \par
When training the \textbf{T+C} variants of our baseline and curriculum models, we first trained the model on the text-only dataset for twenty epochs (instead of eight epochs for image-caption data) and use the same hyperparameter values. We used an NVIDIA A5000 GPU with 24GB vRAM, with half-precision (FP16) to train the models. We provide the total time required to train each model variant in Appendix \ref{sec:comparison_training_times}. For all experiments, we set the random seed to 0 to remove variation in the results due to different random sampling and initialization. We also hold out 5\% of the full image-caption dataset to validate the model. We show the validation loss curves in Appendix \ref{sec:validation_loss_curves}.

\paragraph{Evaluation:} To evaluate the performance of our models, we use the evaluation pipeline provided by challenge \cite{eval-harness,babylm-2024}.
We report the performance of all the variants of the $GIT$ and $Flamingo$ models on the multimodal, and text-based evaluation tasks.

\subsection{Multimodal evaluation datasets}
\label{sec:multimodal_evaluation}
% We evaluate our models on the Winground, VQAv2, and DevBench evaluation datasets.
\paragraph{Winoground}: The Winoground dataset \cite{thrush_and_ross2022winoground} evaluates a model's ability to perform visio-linguistic compositional reasoning. Specifically, given two image-caption pairs, the goal is to match the image to the corresponding caption, where both captions contain an identical set of words, but in a different order (e.g. \textit{It's a fire truck} vs \textit{it's a truck fire}). The dataset consists of 400 examples with 800 unique images and captions. To assess model performance, we use the unpaired text-score metric as provided in the BabyLM evaluation pipeline.

\paragraph{VQAv2:} The VQAv2 dataset \cite{goyal2017makingvvqamatter} is a large-scale visual question answering dataset. It contains open-ended questions about images, requiring models to understand the visual content and generate appropriate answers. We use accuracy as the choice of metric as reported in the BabyLM evaluation pipeline. For this task the model has to select the best answer for a given image and question, in the presence of 7 distractors.
% \textcolor{red}{<TODO: Talk about the 7 distractors provided by the challenge.>}

\paragraph{DevBench:} The DevBench dataset \cite{tan2024devbenchmultimodaldevelopmentalbenchmark} is a multimodal benchmark for developmental evaluation that evaluates how closely a model's outputs align with human responses. It includes tasks such as object recognition, action recognition, and visual question answering, using data from both children and adults. The BabyLM pipeline uses three tasks from the DevBench dataset: (1) The (Lexical) Visual Vocabulary (lex-viz\_vocab) task involves selecting the correct image from several image options based on a given word. (2) The (Grammatical) Test of Reception of Grammar (gram-trog) task involves choosing the correct image based on a sentence, testing grammatical understanding using distractor images that correspond to sentences with different word orderings (e.g. "a white cat sitting on a brown couch" vs. "a brown cat sitting on a white couch"). Finally, (3) the (Semantic) THINGS Similarity (sem-things) task uses Representational Similarity Analysis (RSA) to compare the model’s image similarity judgments with human responses.

\renewcommand{\arraystretch}{1.2} % Adjust this to change vertical spacing.

\begin{table*}[!htb]
    \centering
    \setlength{\tabcolsep}{2pt} % Adjust space between columns
    \renewcommand{\arraystretch}{1.2} % Adjust space between rows
    \begin{tabular}{|p{1.85cm}|p{0.9cm}|p{0.9cm}|p{0.9cm}|p{0.9cm}|p{1.3cm}|p{1.3cm}|p{1cm}|p{0.9cm}|}
        \hline 
        
        \centering \multirow{2}{*}{\textbf{Tasks}} & \multicolumn{2}{c|}{\textbf{$GIT_{Baseline}$}} & \multicolumn{2}{c|}{\textbf{$GIT_{CL}$}} & \multicolumn{2}{c|}{\textbf{$Flamingo_{Baseline}$}} & \multicolumn{2}{c|}{\textbf{$Flamingo_{CL}$}} \\
        \cline{2-9}
          & \centering \textbf{C} & \centering \textbf{T+C} &  \centering \textbf{C} & \centering \textbf{T+C} & \centering \textbf{C} & \centering \textbf{T+C} & \centering \textbf{C} & \centering \textbf{T+C} \cr
          \cline{1-9} 
        \centering Winoground & 54.02 & 55.50 & 51.34 & 55.23 & \centering 50.00 & \centering 51.21 & 51.21 & 50.80 \\ \hline
        \centering VQAv2 & 41.22 & 41.72 & \cellcolor{green}42.84 & \cellcolor{green}43.98 & \centering 41.99 & \centering43.00 & 35.93 & 40.85 \\ \hline

    \end{tabular}
    \caption{Results for baseline and curriculum models on the Winoground and VQAv2 evaluation datasets. \textbf{C}: Model trained on image-caption pairs only (50M words), \textbf{T+C}: the model is first trained on the text-only dataset (20 epochs) and then trained on image-caption pairs (50M+50M=100M words). Green cells: winning variants over corresponding baseline variants. 
}
    \label{tab:multimodal_evaluation}
\end{table*}
% Winoground - accuracy - chance is 50%
% VQAv2 - accuracy (multiple choice) - 1/7 is chance accuracy.
% devbench - 

\begin{table*}[!htb]
    \centering
    \setlength{\tabcolsep}{2pt} % Adjust space between columns
    \renewcommand{\arraystretch}{1.2} % Adjust space between rows
    \begin{tabular}{|p{2.90cm}|p{1cm}|p{1cm}|p{1cm}|p{1cm}|p{1.3cm}|p{1.3cm}|p{1cm}|p{1cm}|}
        \hline 
        
        \centering \multirow{2}{*}{\textbf{Tasks}} & \multicolumn{2}{c|}{\textbf{$GIT_{Baseline}$}} & \multicolumn{2}{c|}{\textbf{$GIT_{CL}$}} & \multicolumn{2}{c|}{\textbf{$Flamingo_{Baseline}$}} & \multicolumn{2}{c|}{\textbf{$Flamingo_{CL}$}} \\
        \cline{2-9}
          & \centering \textbf{C} & \centering \textbf{T+C} &  \centering \textbf{C} & \centering \textbf{T+C} & \centering \textbf{C} & \centering \textbf{T+C} & \centering \textbf{C} & \centering \textbf{T+C} \cr
          \cline{1-9}
        \centering lex-viz\_vocab & 72.27 & 75.63 & \cellcolor{green} 78.15 & 73.11 & \centering 66.39 & \centering 52.94 & 58.82 & \cellcolor{green} 54.62\\ \hline
        \centering gram-trog & 32.89 & 38.16 & 32.29 & \cellcolor{green} 39.47 &\centering 34.21 & \centering 34.21 & 34.21 & \cellcolor{green} 35.53 \\ \hline
        \centering sem-things & 33.39 & 25.79 & 22.83 & \cellcolor{green} 32.08 & \centering 46.46 & \centering 47.99 & \cellcolor{green} 50.21 & \cellcolor{green}51.66\\ \hline \hline
        \centering Avg: $devbench_{acc}$ & 46.18 & 46.52 & 44.63 & \cellcolor{green} 48.22  & \centering 49.02 & \centering 45.05 & 47.75 & \cellcolor{green} 47.27 \\ \hline
    \end{tabular}
    \caption{Accuracy results for baseline and curriculum models on the DevBench dataset. RSA scores are used for sem-things \textbf{C}: Model trained on image-caption pairs only (50M words), \textbf{T+C}: the model is first trained on the text-only dataset (20 epochs) and then trained on image-caption pairs (50M+50M=100M words). Green cells: winning variants over corresponding baseline variants..  
}
    \label{tab:devbench_acc}
\end{table*}

\begin{table*}[!htb]
    \centering
    \setlength{\tabcolsep}{2pt} % Adjust space between columns
    \renewcommand{\arraystretch}{1.2} % Adjust space between rows
    \begin{tabular}{|p{2.90cm}|p{1cm}|p{1cm}|p{1cm}|p{1cm}|p{1.3cm}|p{1.3cm}|p{1cm}|p{1cm}|}
        \hline 
        
        \centering \multirow{2}{*}{\textbf{Tasks}} & \multicolumn{2}{c|}{\textbf{$GIT_{Baseline}$}} & \multicolumn{2}{c|}{\textbf{$GIT_{CL}$}} & \multicolumn{2}{c|}{\textbf{$Flamingo_{Baseline}$}} & \multicolumn{2}{c|}{\textbf{$Flamingo_{CL}$}} \\
        \cline{2-9}
          & \centering \textbf{C} & \centering \textbf{T+C} &  \centering \textbf{C} & \centering \textbf{T+C} & \centering \textbf{C} & \centering \textbf{T+C} & \centering \textbf{C} & \centering \textbf{T+C} \cr
          \cline{1-9}
        \centering lex-viz\_vocab & 68.25 & 68.59 & \cellcolor{green} 70.19 & \cellcolor{green} 70.66 &\centering 64.47 & \centering 57.63 & 63.08 & 57.46\\ \hline
        \centering gram-trog & 44.46 & 46.51 & \cellcolor{green} 44.77 & 45.79 & \centering43.59 & \centering 42.77 &  42.54 & \cellcolor{green} 43.29\\ \hline
        \centering sem-things & 33.39 & 25.79 & 22.83 & \cellcolor{green} 32.08 &\centering 46.46 &\centering 47.99 &\cellcolor{green} 50.21 & \cellcolor{green} 51.66\\ \hline \hline
        \centering Avg: $devbench_{hs}$ & 48.70 & 46.96 & 45.93 &\cellcolor{green}  49.51 & \centering 51.51 & \centering 49.46 & \cellcolor{green} 51.94 & \cellcolor{green} 50.80\\ \hline
    \end{tabular}
    \caption{Human similarity scores for baseline and curriculum models on the DevBench dataset. RSA scores are used for sem-things. \textbf{C}: Model trained on image-caption pairs only (50M words), \textbf{T+C}: the model is first trained on the text-only dataset (20 epochs) and then trained on image-caption pairs (50M+50M=100M words). Green cells: winning variants over corresponding baseline variants.
}
    \label{tab:devbench_hs}
\end{table*}

\subsection{Text-only evaluation datasets}
\label{sec:text_only_data_evaluation}

% We report the model performance on multiple text evaluation datasets. 
% We report our results on the \texttt{BLiMP}, \texttt{BLiMP Supplement}, \texttt{(Super)GLUE}, and \texttt{EWOK} datasets. For \texttt{BLiMP}, \texttt{BLiMP Supplement}, and \texttt{EWOK}, we report zero-shot results, while we fine-tune our models on \texttt{(Super)Glue}.

\paragraph{BLIMP (and BLIMP Supplement):} The \texttt{BLIMP} dataset \cite{warstadt2020blimp} is a benchmark for evaluating syntactic and semantic knowledge in language models. It consists of sentences with systematic variations in syntax and semantics. The \texttt{BLIMP Supplement} extends the original dataset with additional challenging examples.

\paragraph{(Super)GLUE:} The \texttt{(Super)Glue} benchmark \cite{wang2019glue,wang2019superglue} is a collection of diverse natural language understanding tasks designed to evaluate a model's ability to perform well across multiple domains and evaluates generalized linguistic ability. 
The BabyLM challenge includes tasks, \texttt{COLA, SST2, MRPC,
QQP, MNLI, MNLI-MM, QNLI, RTE} from the \texttt{GLUE} benchmark, and the tasks \texttt{BoolQ, RTE and WSC} from \texttt{SuperGLUE} benchmark. To fine tune all our model variants, we use a train batch size of $128$, validation batch size of $16$, learning rate of $5e^{-5}$, early stopping patience of $3$, maximum sequence length of $50$, and maximum number of epochs=10. We used default values for all other hyperparameters provided in the BabyLM evaluation pipeline.

\paragraph{EWOK:} The \texttt{EWOK} dataset \cite{ivanova2024elements} is a zero-shot dataset for evaluating compositional generalization in language models. It consists of sentences with compositional structures that require models to generalize to unseen combinations of words and syntactic patterns.

%We show text-only evaluation results in Table \ref{tab:text_only_evaluation}. 

\section{Results}
\label{sec:results}

As unimodal and multimodal tasks are qualitatively different, we analyze the three experimental variables of interest (curriculum, pretraining \& model type) in the context of each task type. Namely, we report the results for all variants of $GIT$ and $Flamingo$ models across two main task types that differ with respect to their data inputs: (i) multimodal (image+captions), and (i) unimodal (text-only).

\newcommand{\STAB}[1]{\begin{tabular}{@{}c@{}}#1\end{tabular}}

\begin{table*}[!htb]
    \centering
    \setlength{\tabcolsep}{2pt} % Adjust space between columns
    \renewcommand{\arraystretch}{1.2} % Adjust space between rows
    \begin{tabular}{|p{2.8cm}|p{0.9cm}|p{0.9cm}|p{0.9cm}|p{0.9cm}|p{1.3cm}|p{1.3cm}|p{1cm}|p{1cm}|}
        \hline 
        
        \centering \multirow{2}{*}{\textbf{Tasks}} & \multicolumn{2}{c|}{\textbf{$GIT_{Baseline}$}} & \multicolumn{2}{c|}{\textbf{$GIT_{CL}$}} & \multicolumn{2}{c|}{\textbf{$Flamingo_{Baseline}$}} & \multicolumn{2}{c|}{\textbf{$Flamingo_{CL}$}} \\
        \cline{2-9}
          & \centering \textbf{C} & \centering \textbf{T+C} &  \centering \textbf{C} & \centering \textbf{T+C} & \centering \textbf{C} & \centering \textbf{T+C} & \centering \textbf{C} & \centering \textbf{T+C} \cr
          \cline{1-9}
        BliMP Supp & 44.29& 52.89  & \cellcolor{green} 48.61 & 51.24& \centering 44.24 & \centering 52.59 & \cellcolor{green}45.71 & \cellcolor{green} 53.28\\
        \cline{1-9}
        BLiMP filtered & 57.85 & 62.90& \cellcolor{green}61.34 & \cellcolor{green} 64.05 & \centering 57.03 & \centering 59.82 & 55.64 & \cellcolor{green}60.13  \\ \hline
        (Super)$GLUE_{avg}$ & 59.96 & 61.12 & 59.79 & \cellcolor{green} 61.46 & \centering 59.82 & \centering 62.79 & \cellcolor{green} 60.53 & \cellcolor{green} 64.29\\ \hline
        \textbf{$EWOK_{avg}$}  & 50.62 & 51.55 & 49.82 & 50.98 & \centering 50.03 &\centering  50.67 & \cellcolor{green}50.16 & \cellcolor{green}50.71\\ \hline

    \end{tabular}
    \caption{Average results for the text-only evaluation datasets.\textbf{C}: Model trained on image-caption pairs only (50M words), \textbf{T+C}: the model is first trained on the text-only dataset (20 epochs) and then trained on image-caption pairs (50M+50M=100M words). Green cells: winning variants over corresponding baseline variants. }
    \label{tab:avg_text_only_evaluation}
\end{table*}

\subsection{Multimodal (image+captions)}
We show the multimodal evaluations results in Table \ref{tab:multimodal_evaluation} for \texttt{Winoground} and \texttt{VQAv2}, and in Tables  \ref{tab:devbench_acc} (accuracy) and \ref{tab:devbench_hs} (human similarity) for \texttt{DevBench}.

\subsubsection{Curriculum Learning}
The $GIT_{CL}$ model performs better than $GIT_{Baseline}$ on \texttt{VQAv2} and \texttt{DevBench} datasets, with and without pretraining on separate text data. This is not the case for \texttt{Winoground}, which we note has quite unique properties, such as specifically probing model representations for compositionality (see Section \ref{sec:multimodal_evaluation}). \par
We find that $Flamingo_{CL}$ only performs better than its associated baseline ($Flamingo_{Baseline}$) on the \texttt{DevBench} dataset when using accuracy, and when evaluating using human response scores. This result indicates that curriculum training may benefit multimodal model performance when evaluated on benchmark datasets that focus on developmentally plausible evaluation of language models.

\subsubsection{Text Pretraining}
Compared to training on just image-caption data, pretraining with the text-only data (variant \textbf{T+C}) produces higher scores across both $GIT_{Baseline}$ and $GIT_{CL}$ models on \texttt{Winoground} and \texttt{DevBench}, while the results are more mixed for $Flamingo$ models. However, in $Flamingo_{CL}$ on the \texttt{VQAv2} dataset, we see the largest gain in performance due to text pretraining (from 35.93 to 40.85, a gain of $\sim$ 5\% in Table \ref{tab:multimodal_evaluation}). On the \texttt{DevBench} evaluation for $GIT_{CL}$, we also see the 2nd largest gain in performance due to text pretraining (from 44.63 to 48.22 for accuracy, and from 45.93 to 49.51 when using reference human similarity scores; a gain of $\sim$ 4\%). Interestingly, the highest result of all models on the \texttt{Winoground} dataset are the $GIT$ models with text pretraining, suggesting that text-only pretraining is a big contributor to the properties of the \texttt{Winoground} evaluation benchmark (compositionality). However, one must be cautious about generalizing this finding as the performance increase could simply result from the model being trained on more data.

\par As we only use a single seed to report these results, we wanted to confirm that our observation is not simply due to random chance. Thus, we conduct more experiments where we train all $GIT$ variants using two more seeds, and observe a similar pattern in our findings (text pretraining aids model performance). We provide these results in Appendix \ref{sec:git_multimodal_multiple_seeds}.

\subsubsection{Model Type}
The two models differ in their application of attention mechanism and model size, measured by the number of trainable parameters (See Section \ref{sec:framework}). $Flamingo$ has a frozen image encoder (unlike $GIT$) and cross-attention is applied prior to each LM block in the Transformer stack (which internally contains the standard self-attention mechanism). In contrast, $GIT$ uses a projection module to bring image embeddings into the same space as the text embeddings and applies successive self-attention on these vectors. 
% Also, $GIT$'s vision encoder is not frozen. 
We see multiple variants of $GIT$ outperform $Flamingo$ (especially for \texttt{Winoground}, \texttt{VQAv2}, and \texttt{lex-viz\_vocab}, \texttt{gram-trog} subsets for \texttt{DevBench}). In the multimodal evaluation context, we believe this could be due to the ability for $GIT$ to update the parameters of its vision encoder, perhaps additionally by making use of the fact that image tokens can self-attend to one another (unlike the cross-attention in $Flamingo$, which does not have this property).

\subsection{Unimodal (text-only)}
We summarize the results for the unimodal (text-only) evaluation in Table \ref{tab:avg_text_only_evaluation}. This table contains summary results for the three text-only evaluation benchmarks (see Section \ref{sec:text_only_data_evaluation}). Table \ref{tab:text_only_evaluation} contains detailed results on the \texttt{(Super)GLUE} and \texttt{EWOK} benchmarks. We also provide a detailed breakdown of model performance for each text-based task in Appendix \ref{sec:text_only_super_glue_ewok_blimp}.

\subsubsection{Curriculum Learning}
Closely related to the observations for multimodal benchmarks, we see that curriculum learning variants outperform corresponding baselines variants on the unimodal (text-only) benchmarks. Although both $GIT_{CL}$ and $Flamingo_{CL}$ outperformed their corresponding baselines (Tables \ref{tab:avg_text_only_evaluation} and \ref{tab:text_only_evaluation}), the effect was greater in $Flamingo_{CL}$. 
% two of the three amalgamated task results, the highest performing model was trained using curriculum learning (specifically in conjunction with text-only dataset pretraining), suggesting that this interactive effect conveys a benefit to model performance in the unimodal / text-only setting.
\subsubsection{Text Pretraining}
We outline the averaged results in Table \ref{tab:avg_text_only_evaluation} and show that for both $Flamingo$ and $GIT$, text pretraining leads to a gain in performance. In fact, all \textbf{T+C} variants (curriculum and baseline) for both models showe better performance compared to \textbf{C} variants. Coupled with curriculum learning, we observe performance benefits on all text-based evaluation datasets. These results suggest that text pretraining conveys a clear advantage for multimodal models when they are evaluated on certain text-based benchmarks.

% not affect the performance of the models (across both curriculum learning and baselines) for the Ewok tasks. For SuperGLUE, on the other hand, pretraining (and curriculum learning) conveys a clear advantage.  

\subsubsection{Model Type}
Unlike the multimodal results, considering the average results in Table \ref{tab:avg_text_only_evaluation}, there was no consistent pattern where one model type outperformed the other. For example, on \texttt{(Super)GLUE}, both baseline and CL \textbf{T+C} variants of $Flamingo$ outperformed respective $GIT$ variants. However, this was not the case for \texttt{BLIMP filtered}, where we observed the opposite pattern - all variants of $GIT$ outperformed all variants of $Flamingo$. Such a result could result from the fact that both $GIT$ and $Flamingo$ become more similar in their architecture in the text-only evaluation setting. This can stem relaxed requirement to incorporate image information, making both models resemble standard autoregressive Transformer decoders (the trainable parameter count changes in this context because $GIT$'s vision encoder was trainable in the multimodal case, while $Flamingo$'s was frozen). This results in the trainable parameter count for $GIT$ being 198M and 169M for $Flamingo$ (Section \ref{sec:framework}). 

\subsection{Brief Summary of Results}
For the multimodal evaluation, we observe that text pretraining before image-caption training boosts model performance compared to no text pretraing. However, these observations must be cautiously generalized across model types;
% there is no consistent pattern that can generalized across (with the exception of all $Flamingo$ variants on lex-viz\_vocab where we see a decrease in performance - Tables \ref{tab:devbench_acc} and \ref{devbench_hs}, and for $Flamingo_{CL}$ for Winoground, where performance decreases slightly - Table \ref{tab:multimodal_evaluation}).
text pretraining largely conveys a benefit in all $GIT$ models, but this benefit is inconsistent for $Flamingo$. For instance, the $Flamingo_{CL}$ variant benefits from additional text-only pretraining over just image-caption training (for \texttt{VQAv2}, \texttt{gram-trog}, and \texttt{sem-things}), but this effect is unclear for the $Flamingo_{Baseline}$. For $GIT$ model variants, curriculum learning (combined with pretraining) resulted in the best overall model scores on \texttt{VQAv2} and \texttt{DevBench} (considering average scores in Tables \ref{tab:devbench_acc} and \ref{tab:devbench_hs}).

% For the text-only evaluation, we see that the removal of the image component in both the $GIT$ and $Flamingo$ models brings the architectures together in terms of similarity and this likely explains the relatively comparable results across models and tasks. 

For the text-only evaluation, removing the image component from both the $GIT$ and $Flamingo$ models effectively reduces them to text-only transformer architectures with differing number of parameters. This likely explains why the models show similar performance across tasks despite their original multimodal design. Nonetheless, we see that in Table \ref{tab:avg_text_only_evaluation}, the $Flamingo_{CL}$ \textbf{T+C} variant can be more suited to learning representations leading to better scores across the \texttt{SuperGLUE} benchmark, and \texttt{BLiMP supplement} dataset. But on \texttt{BLiMP filtered} (and less pronounced for \texttt{EWOK}), the \textbf{T+C} variant of $GIT_{CL}$ outperforms the \textbf{T+C} variant of $Flamingo_{CL}$.

\section*{Conclusion}
\label{sec:curriculum}
In this study, we explore the application of a curriculum learning (CL) approach to training vision-language models (VLMs) in a limited data setting. We use a custom trained Part-of-Speech (PoS) tagger to determine the complexity of image-caption pairs.
We train two variants for each of the $GIT$ and $Flamingo$ models using curriculum learning and compare their performance against variants trained using standard i.i.d training. 
We find that while CL training shows potential, its benefits are not universally applicable across all $GIT$ and $Flamingo$ variants. However, for certain model configurations, CL enhances performance on a range of downstream, multimodal and text-based tasks (zero-shot and finetuning). Importantly, pretraining VLMs on developmentally plausible text data prior to multimodal training can contribute to performance gains. Nonetheless, generalizing this result requires careful consideration, as factors such as model architecture, training data composition, and the nature of evaluation tasks can significantly affect model performance.

% CL may outperform standard trained models in certain scenarios, such as for a particular architecture, and for specific datasets - this is from a long standing agreement that CL techniques are hard to generalize across multiple architectures and datasets.

\section*{Code and Data Availability}
\label{sec:code}
We release our \href{https://github.com/simpleParadox/baby_lm_2024}{code}, 
model \href{https://osf.io/t9xsh/}{predictions}, and model \href{https://huggingface.co/collections/simpleParadox/babylm-2024-phase-based-curriculum-git-flamingo-66df719cf660074084ec1fee}{checkpoints}.

% Bibliography entries for the entire Anthology, followed by custom entries
%\bibliography{anthology,custom}
% Custom bibliography entries only
\bibliography{acl_latex}

\appendix 

\section{Comparison of Training Times}
\label{sec:comparison_training_times}
We show the comparison of the training times for different baseline and curriculum variants in Table \ref{tab:comparison_training_times}.

\begin{table}[!htb]
\centering
\begin{tabular}{|c|c|c|}
\hline
Model & Variant & Hours \\ \hline
\multirow{2}{*}{$GIT_{Baseline}$} & C & $\sim$ 80  \\ \cline{2-3}
 & T+C & $\sim$ 109 \\ \hline
\multirow{2}{*}{$GIT_{CL}$} & C & $\sim$ 50\\\cline{2-3}
 & T+C & $\sim$ 79 \\ \hline
\multirow{2}{*}{$Flamingo_{Baseline}$} & C & $\sim$ 79\\\cline{2-3}
 & T+C & $\sim$ 105\\ \hline
\multirow{2}{*}{$Flamingo_{CL}$} & C & $\sim$ 46 \\\cline{2-3}
 & T+C &  $\sim$ 72  \\ \hline
\end{tabular}
\caption{Comparison of training times amongst all model variants. These training times include validation loss calculation after every epoch. The pretraining on the text-only dataset (for the T+C variants) accounted for about 29 hours for the $GIT$ model and around 26 hours for the $Flamingo$ model. Curriculum models take fewer hours to train because of the dynamic nature of the training data size that grows during training.}
\label{tab:comparison_training_times}
\end{table}

\section{Validation loss curves}
\label{sec:validation_loss_curves}
\begin{figure}[!htb]
    \centering
    \includegraphics[width=1\linewidth]{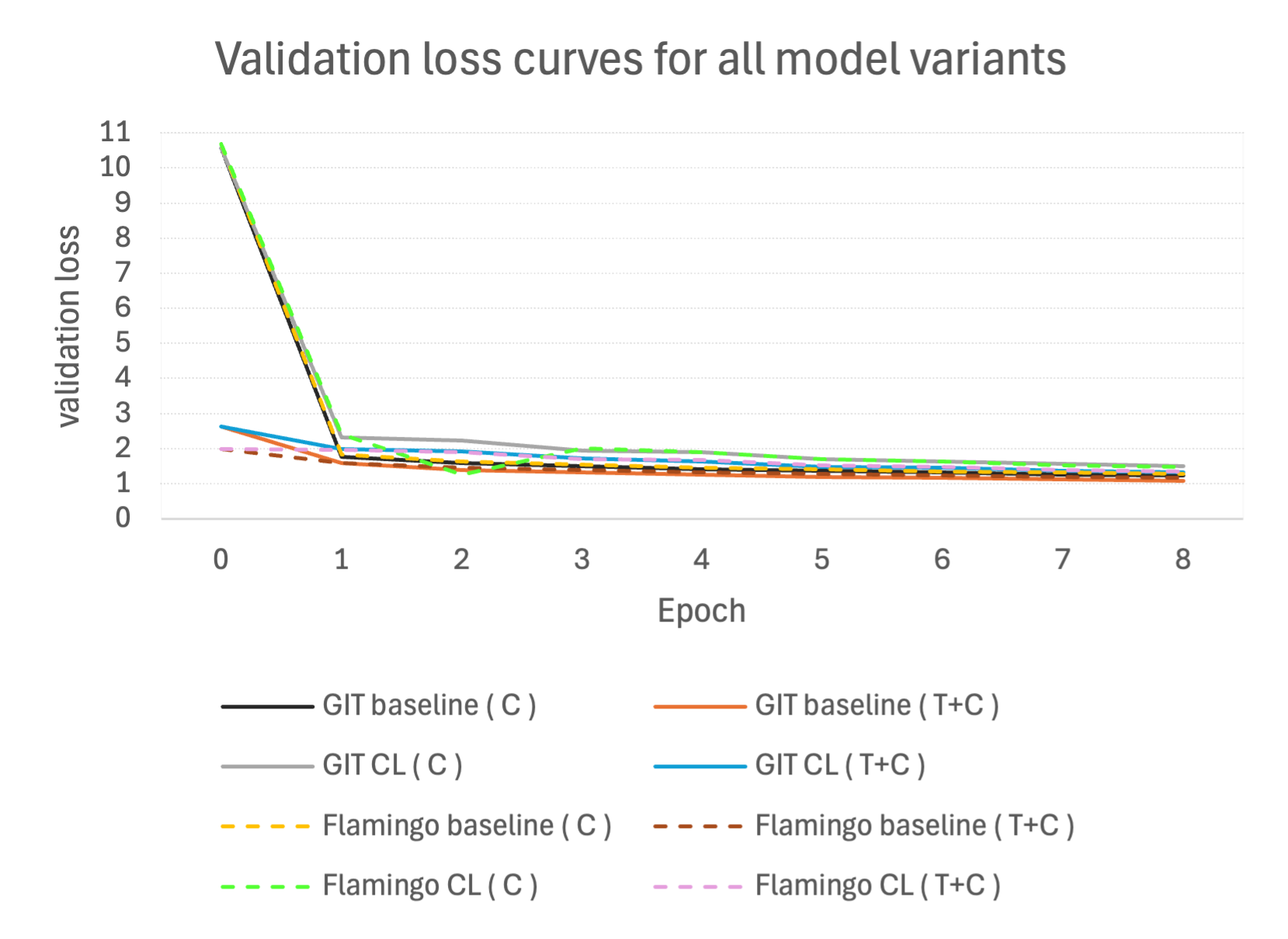}
    \caption{Validation loss curves for all the model variants. $GIT$ variants are shown in solid lines and $Flamingo$ variants are shown in dashed lines. The x-axis denotes the epochs, and the value at the 0th epoch denotes the validation loss of the model before being trained on the image-caption pairs (i.e., before training on the first epoch). For the \textbf{T+C} variants, since the model is pretrained on the text-only dataset before being trained on the image-caption pairs, the loss starts at a lower value compared to the model variants on image-caption data only (\textbf{C}) that were randomly initialized.}
    \label{fig:val_loss_curves_all_variants}
\end{figure}

We show the validation loss curves on a held out 5\% of the image-caption data in Figure \ref{fig:val_loss_curves_all_variants}.
\newpage

\section{GIT model multimodal results across 3 seeds}
\label{sec:git_multimodal_multiple_seeds}
We show the multimodal evaluation results for the different GIT model variants in Tables \ref{tab:git_multimodal_avg_three_seeds} for Winoground and VQAv2, \ref{tab:git_devbench_acc_avg_three_seeds} for accuracy on DevBench, and \ref{tab:git_devbench_hs_avg_three_seeds} human similarity scores on DevBench.

\begin{table}[!htb]
    \centering
    \setlength{\tabcolsep}{2pt} % Adjust space between columns
    \renewcommand{\arraystretch}{1.2} % Adjust space between rows
    \begin{tabular}{|p{1.85cm}|p{0.9cm}|p{0.9cm}|p{0.9cm}|p{0.9cm}|}
        \hline 
        
        \centering \multirow{2}{*}{\textbf{Tasks}} & \multicolumn{2}{c|}{\textbf{$GIT_{Baseline}$}} & \multicolumn{2}{c|}{\textbf{$GIT_{CL}$}}\\
        \cline{2-5}
          & \centering \textbf{C} & \centering \textbf{T+C} &  \centering \textbf{C} & \centering \textbf{T+C} \cr
          \cline{1-5}
        \centering Winoground & 54.02 & 53.71 & 51.52 & \cellcolor{green} 54.65\\ \hline
        \centering VQAv2 & 38.80 & 41.90 & \cellcolor{green} 42.28 & \cellcolor{green} 42.60\\ \hline

    \end{tabular}
    \caption{Results for $GIT$ baseline and $GIT$ curriculum models on the multimodal evaluation datasets averaged across three seeds. \textbf{C}: Model trained on image-caption pairs only (50M words), \textbf{T+C}: the model is first trained on the text-only dataset (20 epochs) and then trained on image-caption pairs (50M+50M=100M words). Green cells: winning variants over corresponding baseline variants.
}
    \label{tab:git_multimodal_avg_three_seeds}
\end{table}

\begin{table}[!htb]
    \centering
    \setlength{\tabcolsep}{2pt} % Adjust space between columns
    \renewcommand{\arraystretch}{1.2} % Adjust space between rows
    \begin{tabular}{|p{2.85cm}|p{0.9cm}|p{0.9cm}|p{0.9cm}|p{0.9cm}|}
        \hline 
        
        \centering \multirow{2}{*}{\textbf{Tasks}} & \multicolumn{2}{c|}{\textbf{$GIT_{Baseline}$}} & \multicolumn{2}{c|}{\textbf{$GIT_{CL}$}}\\
        \cline{2-5}
          & \centering \textbf{C} & \centering \textbf{T+C} &  \centering \textbf{C} & \centering \textbf{T+C} \cr
          \cline{1-5}
        \centering lex-viz\_vocab & 72.93 &	72.55 & \cellcolor{green} 75.91	& 71.71   \\ \hline
        \centering gram-trog & 38.16	& 36.84	& 32.26	& \cellcolor{green} 41.67\\ \hline
        \centering sem-things & 30.88	& 25.61	& 17.34	& \cellcolor{green} 30.78\\ \hline
        \centering $Average_{acc}$ & 47.32	& 45.00	& 41.84	& \cellcolor{green} 48.05 \\ \hline

    \end{tabular}
    \caption{Accuracy results for $GIT$ baseline and $GIT$ curriculum models on the devbench datasets averaged across three seeds. \textbf{C}: Model trained on image-caption pairs only (50M words), \textbf{T+C}: the model is first trained on the text-only dataset (20 epochs) and then trained on image-caption pairs (50M+50M=100M words). Green cells: winning variants over corresponding baseline variants.
}
    \label{tab:git_devbench_acc_avg_three_seeds}
\end{table}

\newcolumntype{P}[1]{>{\centering\arraybackslash}p{#1}}

\begin{table}[!htb]
    \centering
    \setlength{\tabcolsep}{2pt} % Adjust space between columns
    \renewcommand{\arraystretch}{1.2} % Adjust space between rows
    \begin{tabular}{|p{2.85cm}|p{0.9cm}|p{0.9cm}|p{0.9cm}|p{0.9cm}|}
        \hline 
        
        \centering \multirow{2}{*}{\textbf{Tasks}} & \multicolumn{2}{c|}{\textbf{$GIT_{Baseline}$}} & \multicolumn{2}{c|}{\textbf{$GIT_{CL}$}}\\
        \cline{2-5}
          & \centering \textbf{C} & \centering \textbf{T+C} &  \centering \textbf{C} & \centering \textbf{T+C} \cr
          \cline{1-5}
        \centering lex-viz\_vocab & 68.64 &	68.07 & \cellcolor{green} 68.65 &	\cellcolor{green} 68.71 \\ \hline
        \centering gram-trog & 44.90 & 44.61 & 43.72 & \cellcolor{green} 45.71\\ \hline
        \centering sem-things & 30.88	& 25.61	& 17.34	& \cellcolor{green} 30.78\\ \hline
        \centering $Average_{hs}$ & 48.14 &	46.10	& 43.24 &	\cellcolor{green} 48.40 \\ \hline

    \end{tabular}
    \caption{Human similarity results for $GIT$ baseline and $GIT$ curriculum models on the devbench datasets averaged across three seeds. \textbf{C}: Model trained on image-caption pairs only (50M words), \textbf{T+C}: the model is first trained on the text-only dataset (20 epochs) and then trained on image-caption pairs (50M+50M=100M words). Green cells: winning variants over corresponding baseline variants.
}
    \label{tab:git_devbench_hs_avg_three_seeds}
\end{table}

\newpage
\section{Evaluation results on \texttt{(Super)GLUE}, \texttt{EWOK}, and \texttt{BLiMP}}
\label{sec:text_only_super_glue_ewok_blimp}

We show the results for all models and corresponding variants on each individual subtask for the text-only evaluation tasks in Tables \ref{tab:text_only_evaluation} for (Super)GLUE and EWOK, \ref{tab:blimp_supplement_all_subtasks} for BLiMP Supplement, and \ref{tab:blimp_all_subtasks_part_1} , \ref{tab:blimp_all_subtasks_part_2}, \ref{tab:blimp_all_subtasks_part_3}, \ref{tab:blimp_all_subtasks_part_4} for BLiMP.

\begin{table*}[!htb]
    \centering
    \setlength{\tabcolsep}{2pt} % Adjust space between columns
    \renewcommand{\arraystretch}{1.2} % Adjust space between rows
    \begin{tabular}{|p{0.9cm}|p{2.5cm}|p{0.9cm}|p{0.9cm}|p{0.9cm}|p{0.9cm}|P{1.5cm}|P{1cm}|p{1.0cm}|p{1cm}|}
        \hline
        & \centering \multirow{2}{*}{\textbf{Tasks}} & \multicolumn{2}{c|}{\textbf{$GIT_{Baseline}$}} & \multicolumn{2}{c|}{\textbf{$GIT_{CL}$}} & \multicolumn{2}{c|}{\textbf{$Flamingo_{Baseline}$}} & \multicolumn{2}{c|}{\textbf{$Flamingo_{CL}$}} \\
        \cline{3-10} 
        & & \centering \textbf{C} & \centering \textbf{T+C} &  \centering \textbf{C} & \centering \textbf{T+C}  & \centering \textbf{C} & \centering \textbf{T+C} & \centering \textbf{C} & \centering \textbf{T+C}  \cr
        \cline{1-10}
        % Adding the new rows
        % \centering \multirow{2}{*}{\STAB{\rotatebox[origin=c]{90}{BLiMP}}} & BliMP Supp & 44.29& 52.89  & \cellcolor{green} 48.61 & 51.24& 44.24 & 52.59 & 45.71 & \cellcolor{green} 53.28\\
        % \cline{2-10}
        % & BLiMP filtered & 57.85 & 62.90& \cellcolor{green}61.34 & \cellcolor{green} 64.05 & 57.03 & 59.82 & 55.64 & \cellcolor{green}60.13  \\

        % \hline
        % \hline
        \centering \multirow{11}{*}{\STAB{\rotatebox[origin=c]{90}{$SuperGLUE_{ft}$}}} & boolq & 64.04 & 65.2 & 64.04 & \cellcolor{green} 70.21 & 67.77 & 66.91 & \cellcolor{green} 68.07 & 66.54 \\
        \cline{2-10}
        & cola (mcc) & 6.68 & 6.68 & 0.0 & 6.68 & 0.0 & 17.7 & 0.0 & \cellcolor{green} 31.75 \\
        \cline{2-10}
        & mnli & 68.7 & 69.74 & 69.34 & \cellcolor{green}69.93 & 66.24 & 70.03 &\cellcolor{green} 67.07 &\cellcolor{green} 70.37\\
        \cline{2-10}
        & mnli-mm & 69.43 & 70.22 & 69.26 & \cellcolor{green} 70.77 & 66.9 & 70.2 & 66.35 & \cellcolor{green}71.42\\
        \cline{2-10}
        & mrpc (f1) & 82.12 & 82.13 & 81.23 & 81.35 & 81.05 & 82.51 & 79.87 & 82.39\\
        \cline{2-10}
        & multirc & 55.57 &  57.43 & \cellcolor{green} 57.55 & 56.97 & 60.81 & 53.55 & 58.21 & \cellcolor{green} 56.23 \\
        \cline{2-10}
        & qnli & 63.14 & 64.42 & \cellcolor{green}67.5 & \cellcolor{green}65.59 & 65.81 & 68.92 & \cellcolor{green} 67.86 & \cellcolor{green}69.91 \\
        \cline{2-10}
        & qqp (f1)& 80.92 & 81.7 & 80.12 & 81.53 & 79.83 & 82.05 & \cellcolor{green}79.91 & 81.88 \\
        \cline{2-10}
        & rte& 46.04 & 48.92 & 46.04 & 46.04 & 46.04 & 52.52 & \cellcolor{green} 56.12 & 46.04 \\
        \cline{2-10}
        & sst2& 84.40 & 87.39 & 84.17 &\cellcolor{green} 88.53 & 85.09 & 87.84 & 83.94 & \cellcolor{green}88.30 \\
        \cline{2-10}
        & wsc & 38.46 & 38.46 & 38.46 & 38.46 & 38.46 & 38.46 & 38.46 & \cellcolor{green} 42.31 \\
        \hline
        \hline
        \centering \multirow{12}{*}{\STAB{\rotatebox[origin=c]{90}{EWOK}}} 
        % \cline{2-10}
        & agent prop & 50.05 & 50.14 & 50.09 & 49.59 & 49.46 & 50.32 & \cellcolor{green}49.91 & 49.68\\
        \cline{2-10}
        & mat-dynam & 51.56 & 52.21 & 51.30 & 50.65 & 49.48 & 52.21 & \cellcolor{green} 50.52 & \cellcolor{green}54.42\\
        \cline{2-10}
        & mat-prop  & 50.59 & 52.35 & 47.06 & 49.41 & 46.47 & 53.53 &\cellcolor{green} 51.76 & 51.18\\
        \cline{2-10}
        & phy-dynam & 49.17 &  55.83 & 48.33 & \cellcolor{green}58.33 & 54.17 & 48.33 & 50.0 & \cellcolor{green}51.67\\
        \cline{2-10}
        & phy-inter & 49.64 & 50.0 &\cellcolor{green} 50.18 & \cellcolor{green}50.18 & 50.18 & 49.1 & 48.74 & 49.1\\
        \cline{2-10}
        & phy-relation & 50.24 & 49.88 & \cellcolor{green} 50.61 & 49.51 & 52.57 & 50.12 & 49.27 & \cellcolor{green}50.86\\
        \cline{2-10}
        & quant-prop & 51.91 & 50.96 & 49.68 & 50.96 & 49.36 & 53.5 & \cellcolor{green}50.64 & 50.0\\
        \cline{2-10}
        & social-interac & 50.34 & 50.34 & 50.34 & 49.66 & 49.32 & 49.32 & \cellcolor{green}49.66 & \cellcolor{green}50.0\\
        \cline{2-10}
        & social-prop &  50.3 & 50.91 & \cellcolor{green}50.91 & 50.0 & 50.0 & 49.09 & \cellcolor{green}50.61 & \cellcolor{green}50.0\\
        \cline{2-10}
        & social-relation & 50.32 & 51.42 & 49.94 & 50.0 & 49.29 & 50.0 & \cellcolor{green}50.45 &\cellcolor{green} 50.06\\
        \cline{2-10}
        & spatial-relation & 52.65 & 53.06 & 49.59 & 52.45 & 50.0 & 51.84 & \cellcolor{green} 50.20 & 50.82\\
        \hline

    \end{tabular}
    \caption{Breakdown of model performance on each subtask for the\texttt{(Super)Glue} and \texttt{EWOK} datasets. Cells highlighted in Green denote winning variants compared to corresponding baseline variants. }
    \label{tab:text_only_evaluation}
\end{table*}

\begin{table*}[!htb]
    \centering
    \setlength{\tabcolsep}{2pt} % Adjust space between columns
    \renewcommand{\arraystretch}{1.2} % Adjust space between rows
    \begin{tabular}{|p{0.9cm}|p{3.5cm}|p{0.9cm}|p{0.9cm}|p{0.9cm}|p{0.9cm}|p{1.5cm}|p{0.9cm}|p{1.0cm}|p{1cm}|}
        \hline
        & \centering \multirow{2}{*}{\textbf{Tasks}} & \multicolumn{2}{c|}{\textbf{$GIT_{Baseline}$}} & \multicolumn{2}{c|}{\textbf{$GIT_{CL}$}} & \multicolumn{2}{c|}{\textbf{$Flamingo_{Baseline}$}} & \multicolumn{2}{c|}{\textbf{$Flamingo_{CL}$}} \\
        \cline{3-10} 
        & & \centering \textbf{C} & \centering \textbf{T+C} &  \centering \textbf{C} & \centering \textbf{T+C}  & \centering \textbf{C} & \centering \textbf{T+C} & \centering \textbf{C} & \centering \textbf{T+C} \cr
        \cline{1-10}

        \centering \multirow{6}{*}{\STAB{\rotatebox[origin=c]{90}{BLiMP Supplement}}} 
           &  hypernym & 47.86 & 48.81 & \cellcolor{green} 49.76 & \cellcolor{green} 48.93 & 49.17 & 48.93 & 48.1 &\cellcolor{green} 51.19\\ \cline{2-10}
        &  qa\_congruence\_easy & 29.69 & 51.56 & \cellcolor{green}35.94 & 50.0 & 32.81 & 51.56 &\cellcolor{green} 37.5 & \cellcolor{green}53.12\\ \cline{2-10}
        &  qa\_congruence\_tricky & 27.88 & 24.24 & 27.27 & 20.0 & 20.0 & 30.91 & \cellcolor{green} 27.27 & 28.48\\ \cline{2-10}
        &  subject\_aux\_inversion & 66.02 & 83.76 &\cellcolor{green} 80.06 & 83.68 & 68.53 & 81.54 & \cellcolor{green} 71.4 &\cellcolor{green} 82.91\\ \cline{2-10}
        &  turn\_taking & 50.0 & 56.07 & 50.0 & 53.57 & 50.71 & 50.0 & 44.29 & \cellcolor{green}50.71\\ \cline{2-10}
        &  \textbf{Average} & 44.29 & 52.89 &\cellcolor{green} 48.61 & 51.24 & 44.24 & 52.59 & \cellcolor{green}45.71 & \cellcolor{green} 53.28\\
        \hline

    \end{tabular}
    \caption{Breakdown of model performance on each subtask for the \texttt{BLiMP Supplement} dataset. Cells highlighted in green denote winning variants compared to corresponding baseline variants. }
    \label{tab:blimp_supplement_all_subtasks}
\end{table*}

\begin{table*}[!htb]
    \centering
    \setlength{\tabcolsep}{2pt} % Adjust space between columns
    \renewcommand{\arraystretch}{1.2} % Adjust space between rows
    \begin{tabular}{|p{0.9cm}|p{3.5cm}|p{0.9cm}|p{0.9cm}|p{0.9cm}|p{0.9cm}|P{1.5cm}|P{1cm}|p{1.0cm}|p{1cm}|}
        \hline
        & \centering \multirow{2}{*}{\textbf{Tasks}} & \multicolumn{2}{c|}{\textbf{$GIT_{Baseline}$}} & \multicolumn{2}{c|}{\textbf{$GIT_{CL}$}} & \multicolumn{2}{c|}{\textbf{$Flamingo_{Baseline}$}} & \multicolumn{2}{c|}{\textbf{$Flamingo_{CL}$}} \\
        \cline{3-10} 
        & & \centering \textbf{C} & \centering \textbf{T+C} &  \centering \textbf{C} & \centering \textbf{T+C}  & \centering \textbf{C} & \centering \textbf{T+C} & \centering \textbf{C} & \centering \textbf{T+C} \cr
        \cline{1-10}

        \centering \multirow{30}{*}{\STAB{\rotatebox[origin=c]{90}{BLiMP}}}
&  determiner\_noun \_agreement\_with\_adj \_irregular\_1 & 64.62 & 74.51 & \cellcolor{green} 71.87 & \cellcolor{green} 76.32 & 50.56 & 62.53 & 49.86 & \cellcolor{green}67.69\\ \cline{2-10}
&  principle\_A\_domain\_3 & 51.75 & 51.97 & 48.67 & 51.22 & 48.46 & 48.57 &\cellcolor{green} 49.31 & 45.59\\ \cline{2-10}
&  sentential\_negation \_npi\_scope & 47.65 & 61.31 & \cellcolor{green}57.52 & 55.57 & 56.83 & 54.54 & 55.57 & 50.86\\ \cline{2-10}
&  complex\_NP\_island & 41.13 & 51.89 & \cellcolor{green}41.61 & \cellcolor{green}54.37 & 58.87 & 43.5 & \cellcolor{green}62.17 & 41.13\\ \cline{2-10}
&  irregular\_plural \_subject\_verb \_agreement\_1 & 55.35 & 64.68 & \cellcolor{green} 63.06 & 64.18 & 49.5 & 57.71 & \cellcolor{green}51.87 & \cellcolor{green}60.45\\ \cline{2-10}
&  distractor\_agreement \_relational\_noun & 41.62 & 46.7 & \cellcolor{green} 47.21 & \cellcolor{green} 51.27 & 52.03 & 46.83 & 49.37 & \cellcolor{green}47.46\\ \cline{2-10}
&  matrix\_question \_npi\_licensor\_present & 3.98 & 44.78 & 4.2 & 33.05 & 84.5 & 59.85 & 35.84 & 39.5\\ \cline{2-10}
&  passive\_2 & 70.65 & 70.32 & \cellcolor{green}72.54 &\cellcolor{green} 72.2 & 70.32 & 70.1 & \cellcolor{green}72.09 & 64.12\\ \cline{2-10}
&  adjunct\_island & 78.45 & 64.12 & 48.38 & \cellcolor{green}66.38 & 55.6 & 59.81 & \cellcolor{green}63.25 & 56.03\\ \cline{2-10}
&  wh\_vs\_that\_with\_gap & 16.1 & 26.55 & 8.05 & 25.9 & 12.19 & 14.47 & \cellcolor{green}35.8 & \cellcolor{green}17.74\\ \cline{2-10}
&  irregular\_past \_participle\_adjectives & 59.63 & 66.6 & \cellcolor{green}79.19 & 63.68 & 46.51 & 48.8 & 45.37 & \cellcolor{green}67.01\\ \cline{2-10}
&  drop\_argument & 71.96 & 74.02 &\cellcolor{green} 73.8 &\cellcolor{green} 76.41 & 70.87 & 70.0 & 70.11 & 68.91\\ \cline{2-10}
&  principle\_A\_domain\_2 & 49.62 & 57.7 &\cellcolor{green} 57.16 & \cellcolor{green}59.02 & 46.34 & 50.93 &\cellcolor{green} 50.82 &\cellcolor{green} 56.28\\ \cline{2-10}
&  anaphor\_gender \_agreement & 45.21 & 46.04 & 36.77 & \cellcolor{green}47.79 & 74.46 & 47.79 & 42.33 & 39.55\\ \cline{2-10}
&  wh\_questions\_subject \_gap\_long\_distance & 93.0 & 85.53 & 97.9 &\cellcolor{green} 89.5 & 81.68 & 88.8 & 61.38 & \cellcolor{green}89.96\\ \cline{2-10}
&  only\_npi\_licensor \_present & 61.68 & 74.72 & \cellcolor{green}93.99 & 52.72 & 72.22 & 92.52 & \cellcolor{green} 97.05 & 58.28\\ \cline{2-10}
&  intransitive & 54.84 & 60.02 & 53.57 &\cellcolor{green} 61.98 & 57.49 & 59.1 & \cellcolor{green}57.6 & \cellcolor{green}60.14\\ \cline{2-10}
&  ellipsis\_n\_bar\_1 & 43.64 & 49.88 & \cellcolor{green}52.37 & \cellcolor{green}59.6 & 38.4 & 61.97 & \cellcolor{green}51.0 & 52.12\\ \cline{2-10}
&  regular\_plural\_subject \_verb\_agreement\_1 & 44.16 & 58.54 & \cellcolor{green}53.15 & \cellcolor{green}58.76 & 49.44 & 55.84 & 45.96 &\cellcolor{green} 61.12\\ \cline{2-10}
&  principle\_A\_domain\_1 & 84.57 & 93.0 &\cellcolor{green} 96.83 & 91.79 & 57.99 & 93.0 & \cellcolor{green}93.22 & 80.42\\ \cline{2-10}
&  irregular\_past \_participle\_verbs & 63.8 & 65.39 & 58.6 & 59.45 & 61.04 & 66.56 & 49.26 & \cellcolor{green}68.05\\ \cline{2-10}
&  sentential\_subject \_island & 54.63 & 62.12 & \cellcolor{green}67.33 & 56.71 & 53.69 & 51.93 & 49.84 & \cellcolor{green}63.89\\ \hline

   \end{tabular}
    \caption{\texttt{BLIMP} - individual task results. Cells highlighted in Green denote winning variants compared to corresponding baseline variants. }
    \label{tab:blimp_all_subtasks_part_1}
\end{table*}

\begin{table*}[!htb]
    \centering
    \setlength{\tabcolsep}{2pt} % Adjust space between columns
    \renewcommand{\arraystretch}{1.2} % Adjust space between rows
    \begin{tabular}{|p{0.9cm}|p{3.5cm}|p{0.9cm}|p{0.9cm}|p{0.9cm}|p{0.9cm}|P{1.5cm}|P{1cm}|p{1.0cm}|p{1cm}|}
        \hline
        & \centering \multirow{2}{*}{\textbf{Tasks}} & \multicolumn{2}{c|}{\textbf{$GIT_{Baseline}$}} & \multicolumn{2}{c|}{\textbf{$GIT_{CL}$}} & \multicolumn{2}{c|}{\textbf{$Flamingo_{Baseline}$}} & \multicolumn{2}{c|}{\textbf{$Flamingo_{CL}$}} \\
        \cline{3-10} 
        & & \centering \textbf{C} & \centering \textbf{T+C} &  \centering \textbf{C} & \centering \textbf{T+C}  & \centering \textbf{C} & \centering \textbf{T+C} & \centering \textbf{C} & \centering \textbf{T+C} \cr
        \cline{1-10}

        \centering \multirow{20}{*}{\STAB{\rotatebox[origin=c]{90}{BLiMP}}}
&  wh\_vs\_that\_with\_gap \_long\_distance & 13.52 & 10.0 & 5.49 & \cellcolor{green}10.22 & 15.6 & 8.46 & \cellcolor{green}40.77 & \cellcolor{green}12.2\\ \cline{2-10}
&  principle\_A\_recons-truction & 54.6 & 50.36 & 53.05 & 35.26 & 56.05 & 53.26 & 50.47 &\cellcolor{green} 55.43\\ \cline{2-10}
&  regular\_plural\_subject\_ verb \_agreement\_2 & 55.03 & 66.88 & \cellcolor{green}64.44 &\cellcolor{green} 68.25 & 48.99 & 51.43 & \cellcolor{green}51.22 & \cellcolor{green}61.9\\ \cline{2-10}
&  ellipsis\_n\_bar\_2 & 29.59 & 51.93 &\cellcolor{green} 31.76 & \cellcolor{green}53.26 & 37.92 & 45.41 &\cellcolor{green} 33.57 & \cellcolor{green}55.68\\ \cline{2-10}
&  determiner\_noun \_agreement\_with \_adj\_irregular\_2 & 65.36 & 75.71 & \cellcolor{green}70.12 & \cellcolor{green}77.5 & 60.0 & 65.12 & 57.26 & \cellcolor{green}68.33\\ \cline{2-10}
&  passive\_1 & 78.1 & 71.55 & \cellcolor{green}80.48 & \cellcolor{green}76.19 & 70.36 & 75.83 &\cellcolor{green} 77.02 & 71.9\\ \cline{2-10}
&  irregular\_plural \_subject\_verb \_agreement\_2 & 59.64 & 68.61 &\cellcolor{green} 71.86 & \cellcolor{green}67.94 & 48.88 & 60.09 & \cellcolor{green}55.83 & \cellcolor{green}69.06\\ \cline{2-10}
&  existential\_there \_subject\_raising & 54.11 & 75.65 & \cellcolor{green}56.06 &\cellcolor{green} 77.81 & 59.74 & 67.42 & 55.3 &\cellcolor{green} 71.21\\ \cline{2-10}
&  left\_branch\_island \_echo\_question & 52.69 & 18.69 & \cellcolor{green}61.14 & 18.27 & 22.39 & 23.34 & 6.65 &\cellcolor{green} 33.37\\ \cline{2-10}
&  expletive\_it\_object \_raising & 63.9 & 63.77 & 62.32 & 62.45 & 63.37 & 64.16 & 61.92 & 63.77\\ \cline{2-10}
&  coordinate\_structure \_constraint\_object \_extraction & 36.14 & 33.4 & \cellcolor{green}51.74 & \cellcolor{green}53.74 & 40.99 & 50.26 & \cellcolor{green}46.36 & \cellcolor{green}61.54\\ \cline{2-10}
&  causative & 58.07 & 67.48 & 56.48 & \cellcolor{green}70.17 & 52.57 & 60.15 & 50.12 & 59.78\\ \cline{2-10}
&  npi\_present\_2 & 38.4 & 61.38 & \cellcolor{green}45.19 & 58.64 & 46.28 & 61.6 & 26.15 & 44.64\\ \cline{2-10} \hline
   \end{tabular}
    \caption{BLIMP - individual task results continued. Cells highlighted in Green denote winning variants compared to corresponding baseline variants. }
    \label{tab:blimp_all_subtasks_part_2}
\end{table*}

% \newcolumntype{L}{>{\centering\arraybackslash}m{3cm}}
\begin{table*}[!htb]
    \centering
    \setlength{\tabcolsep}{2pt} % Adjust space between columns
    \renewcommand{\arraystretch}{1.2} % Adjust space between rows
    \begin{tabular}{|p{0.9cm}|p{4cm}|p{0.9cm}|p{0.9cm}|p{0.9cm}|p{0.9cm}|P{1.5cm}|P{0.9cm}|p{1.0cm}|p{1cm}|}
        \hline
        & \centering \multirow{2}{*}{\textbf{Tasks}} & \multicolumn{2}{c|}{\textbf{$GIT_{Baseline}$}} & \multicolumn{2}{c|}{\textbf{$GIT_{CL}$}} & \multicolumn{2}{c|}{\textbf{$Flamingo_{Baseline}$}} & \multicolumn{2}{c|}{\textbf{$Flamingo_{CL}$}} \\
        \cline{3-10} 
        & & \centering \textbf{C} & \centering \textbf{T+C} &  \centering \textbf{C} & \centering \textbf{T+C}  & \centering \textbf{C} & \centering \textbf{T+C} & \centering \textbf{C} & \centering \textbf{T+C} \cr
        \cline{1-10}

        \centering \multirow{30}{*}{\STAB{\rotatebox[origin=c]{90}{BLiMP}}}
&  animate\_subject\_trans & 46.05 & 44.53 & 22.64 & 38.68 & 30.55 & 49.84 & \cellcolor{green}64.46 &\cellcolor{green} 66.31\\ \cline{2-10}
&  transitive & 69.93 & 73.04 & \cellcolor{green}71.08 &\cellcolor{green} 75.23 & 52.65 & 63.59 &\cellcolor{green} 60.25 & 58.99\\ \cline{2-10}
&  determiner\_noun \_agreement\_with\_adj\_2 & 65.99 & 78.53 & 65.57 &\cellcolor{green} 81.62 & 50.05 & 60.04 &\cellcolor{green} 56.11 & \cellcolor{green}70.24\\ \cline{2-10}
&  determiner\_noun \_agreement\_irregular\_2 & 75.12 & 81.34 & 72.2 & \cellcolor{green}84.88 & 63.17 & 73.78 & 61.71 &\cellcolor{green} 77.56\\ \cline{2-10}
&  left\_branch\_island \_simple\_question & 46.37 & 36.8 & \cellcolor{green}62.78 & 35.44 & 39.54 & 45.53 & 33.96 & 37.64\\ \cline{2-10}
&  wh\_vs\_that\_no\_gap & 85.13 & 91.17 & \cellcolor{green}94.19 &\cellcolor{green} 94.89 & 90.36 & 93.26 & 64.0 & 93.26\\ \cline{2-10}
&  tough\_vs\_raising\_2 & 67.72 & 69.24 &\cellcolor{green} 74.57 & \cellcolor{green}72.5 & 51.74 & 63.7 & \cellcolor{green}56.85 & \cellcolor{green}72.07\\ \cline{2-10}
&  principle\_A\_case\_1 & 99.78 & 100.0 & 99.78 & 100.0 & 93.31 & 98.79 &\cellcolor{green} 98.25 & 98.03\\ \cline{2-10}
&  wh\_questions\_subject\_gap & 81.51 & 85.41 & \cellcolor{green}91.43 & \cellcolor{green}88.86 & 82.63 & 89.2 & 72.16 & 87.53\\ \cline{2-10}
&  only\_npi\_scope & 35.72 & 50.3 & \cellcolor{green}69.3 & 46.12 & 79.81 & 61.05 & 75.03 & 39.67\\ \cline{2-10}
&  distractor\_agreement \_relative\_clause & 43.51 & 46.73 & 40.07 & 44.78 & 54.31 & 48.91 & 53.16 & 48.56\\ \cline{2-10}
&  existential\_there \_quantifiers\_2 & 58.29 & 17.34 & 38.31 & \cellcolor{green}30.63 & 19.87 & 34.03 &\cellcolor{green} 21.08 & 18.33\\ \cline{2-10}
&  determiner\_noun \_agreement\_1 & 74.27 & 81.92 & 71.69 & \cellcolor{green} 84.39 & 56.51 & 70.72 & \cellcolor{green}58.56 & \cellcolor{green}75.03\\ \cline{2-10}
&  superlative\_quantifiers\_1 & 61.08 & 71.71 & 48.52 & \cellcolor{green}85.39 & 51.17 & 39.43 & \cellcolor{green}57.3 & 37.59\\ \cline{2-10}
&  determiner\_noun \_agreement\_with\_adjective\_1 & 64.84 & 80.49 &\cellcolor{green} 69.77 &\cellcolor{green} 81.89 & 56.81 & 63.88 &\cellcolor{green} 57.56 & \cellcolor{green}71.28\\ \cline{2-10}
&  sentential\_negation \_npi\_licensor\_present & 90.64 & 99.35 & \cellcolor{green}99.56 & 92.49 & 91.95 & 99.56 & 72.91 & 98.91\\ \cline{2-10}
&  wh\_questions\_object\_gap & 55.65 & 49.71 &\cellcolor{green} 73.69 & \cellcolor{green}57.97 & 73.11 & 64.96 & 72.53 & 60.3\\ \cline{2-10}
&  determiner\_noun \_agreement\_2 & 69.92 & 80.88 & \cellcolor{green}71.21 & \cellcolor{green}82.92 & 52.52 & 66.38 &\cellcolor{green} 57.14 &\cellcolor{green} 75.94\\ \cline{2-10}
&  existential\_there \_quantifiers\_1 & 78.06 & 92.15 & 77.96 & \cellcolor{green}94.52 & 75.48 & 66.77 & 74.73 & \cellcolor{green} 68.6\\ \cline{2-10}
&  inchoative & 43.04 & 50.53 & 40.12 & \cellcolor{green}52.16 & 43.63 & 49.01 & \cellcolor{green}44.91 & \cellcolor{green}50.76\\ \cline{2-10}
&  coordinate\_structure \_constraint\_complex\_left \_branch & 40.07 & 30.13 & \cellcolor{green}55.08 & 27.37 & 35.76 & 38.41 & 33.11 & 30.13\\ \cline{2-10}
&  superlative\_quantifiers\_2 & 86.51 & 75.56 & \cellcolor{green}88.03 & \cellcolor{green}79.11 & 78.19 & 48.68 & 76.27 & 46.96\\ \cline{2-10}
&  npi\_present\_1 & 40.48 & 52.59 & \cellcolor{green}53.14 &\cellcolor{green} 57.43 & 48.4 & 57.98 & \cellcolor{green}50.72 & 57.87\\ \cline{2-10}
&  wh\_island & 17.71 & 27.92 & \cellcolor{green}32.08 &\cellcolor{green} 51.88 & 61.25 & 18.12 & 48.75 & \cellcolor{green}40.42\\ \cline{2-10}
&  existential\_there\_object \_raising & 70.44 & 66.13 & 67.73 & 60.96 & 68.23 & 70.94 & 66.26 & 67.98\\\hline

    \end{tabular}
    \caption{BLIMP - individual task results continued. Cells highlighted in Green denote winning variants compared to corresponding baseline variants. }
    \label{tab:blimp_all_subtasks_part_3}
\end{table*}

\begin{table*}[!htb]
    \centering
    \setlength{\tabcolsep}{2pt} % Adjust space between columns
    \renewcommand{\arraystretch}{1.2} % Adjust space between rows
    \begin{tabular}{|p{0.9cm}|p{4cm}|p{0.9cm}|p{0.9cm}|p{0.9cm}|p{0.9cm}|P{1.5cm}|P{0.9cm}|p{1.0cm}|p{1cm}|}
        \hline
        & \centering \multirow{2}{*}{\textbf{Tasks}} & \multicolumn{2}{c|}{\textbf{$GIT_{Baseline}$}} & \multicolumn{2}{c|}{\textbf{$GIT_{CL}$}} & \multicolumn{2}{c|}{\textbf{$Flamingo_{Baseline}$}} & \multicolumn{2}{c|}{\textbf{$Flamingo_{CL}$}} \\
        \cline{3-10} 
        & & \centering \textbf{C} & \centering \textbf{T+C} &  \centering \textbf{C} & \centering \textbf{T+C}  & \centering \textbf{C} & \centering \textbf{T+C} & \centering \textbf{C} & \centering \textbf{T+C} \cr
        \cline{1-10}

        \centering \multirow{9}{*}{\STAB{\rotatebox[origin=c]{90}{BLiMP}}}

&  wh\_vs\_that\_no\_gap\_long \_distance & 86.4 & 94.4 &\cellcolor{green} 94.97 & \cellcolor{green}96.57 & 89.6 & 94.29 & 61.37 & 93.37\\ \cline{2-10}
&  principle\_A\_c\_command & 69.13 & 71.88 & 66.07 &\cellcolor{green} 75.58 & 57.61 & 75.69 & \cellcolor{green}66.17 & \cellcolor{green}78.12\\ \cline{2-10}
&  animate\_subject\_passive & 61.45 & 70.28 &\cellcolor{green} 73.85 & \cellcolor{green}72.18 & 63.13 & 65.14 & 60.67 & \cellcolor{green}72.51\\ \cline{2-10}
&  anaphor\_number\_agre-ement & 73.15 & 80.34 & 62.41 & \cellcolor{green}86.14 & 71.0 & 72.82 & 49.41 &\cellcolor{green} 74.22\\ \cline{2-10}
&  determiner\_noun \_agreement\_irregular\_1 & 64.61 & 70.63 & \cellcolor{green}67.25 & \cellcolor{green}75.18 & 59.47 & 62.56 & 54.63 & \cellcolor{green}73.57\\ \cline{2-10}
&  tough\_vs\_raising\_1 & 33.12 & 49.89 & 28.69 & 46.62 & 51.9 & 46.41 & 47.36 & 39.45\\ \cline{2-10}
&  principle\_A\_case\_2 & 62.84 & 77.27 & \cellcolor{green}72.35 & \cellcolor{green}79.23 & 54.97 & 62.95 & 48.96 & 62.62\\ \hline
% &  \textbf{Average} & 57.85 & 62.90 & \cellcolor{green} 61.34 & \cellcolor{green}64.05 & 57.03 & 59.82 & 55.64 & \cellcolor{green}60.13\\ \hline

\end{tabular}
    \caption{BLIMP - individual task results continued. Cells highlighted in Green denote winning variants compared to corresponding baseline variants. }
    \label{tab:blimp_all_subtasks_part_4}
\end{table*}

\end{document}